\documentclass[lettersize,journal]{IEEEtran}
\usepackage{amsmath,amsfonts}
\usepackage{algorithmic}
\usepackage{algorithm}
\usepackage{array}
\usepackage[caption=false,font=normalsize,labelfont=sf,textfont=sf]{subfig}
\usepackage{textcomp}
\usepackage{stfloats}
\usepackage{url}
\usepackage{verbatim}
\usepackage{graphicx}
\usepackage{cite}
\usepackage{longtable}
\usepackage{booktabs}
\usepackage{array}
\usepackage{cuted} 

\usepackage[html]{xcolor}
\usepackage{graphicx}
\usepackage{comment}
\usepackage{amsmath,amssymb,amsfonts}
\usepackage{algorithmic}
\usepackage{textcomp}
\usepackage{picinpar}
\usepackage{flushend}
\usepackage{colortbl}
\usepackage{color}
\usepackage{soul}
\usepackage{multirow}
\usepackage{pifont}
\usepackage{alltt}
\usepackage{enumerate}
\usepackage{pbox}

\begin{document}

\title{Learning and Generating Diverse Residential Load Patterns Using GAN with Weakly-Supervised Training and Weight Selection}

\author{Xinyu~Liang and~Hao~Wang,~\IEEEmembership{Member,~IEEE}
\thanks{This work was supported in part by the Australian Research Council (ARC) Discovery Early Career Researcher Award (DECRA) under Grant DE230100046. (Corresponding author: Hao Wang.)}
\thanks{X. Liang and H. Wang are with Department of Data Science and AI, Faculty of Information Technology and Monash Energy Institute, Monash University, Melbourne, VIC 3800, Australia (e-mails: \{xinyu.liang, hao.wang2\}@monash.edu).}
}

\maketitle

\begin{abstract}
The scarcity of high-quality residential load data can pose obstacles for decarbonizing the residential sector as well as effective grid planning and operation. The above challenges have motivated research into generating synthetic load data, but existing methods faced limitations in terms of scalability, diversity, and similarity. This paper proposes a Generative Adversarial Network-based Synthetic Residential Load Pattern (RLP-GAN) generation model, a novel weakly-supervised GAN framework, leveraging an over-complete autoencoder to capture dependencies within complex and diverse load patterns and learn household-level data distribution at scale. We incorporate a model weight selection method to address the mode collapse problem and generate load patterns with high diversity. We develop a holistic evaluation method to validate the effectiveness of RLP-GAN using real-world data of 417 households. The results demonstrate that RLP-GAN outperforms state-of-the-art models in capturing temporal dependencies and generating load patterns with higher similarity to real data. Furthermore, we have publicly released the RLP-GAN generated synthetic dataset\footnote{https://github.com/AdamLiang42/SynResLoadPattern}, which comprises one million synthetic residential load pattern profiles.
\end{abstract}


\newcommand{\dou}[1]{\textcolor{red}{\textsc{Changed:} #1}}

\section{Introduction}
\IEEEPARstart{T}{he} residential sector is a substantial contributor to the energy consumption, accounting for around 21\% percent of overall energy consumption in the United States \cite{decarolis2023annual} and 30\% in the European Union as of 2020 \cite{statistics2020energy}. 
Residential load data are crucial for research in load profiling, demand response, and home energy management to improve energy efficiency, promote renewable energy, and optimize grid infrastructure, thereby reducing carbon emissions and contributing to climate change mitigation \cite{tang2022machine,arpogaus2023short}. Integrated with information and communications technologies, residential load data can support real-time pricing models and inform load shifting, offering financial savings for consumers while balancing loads for electricity providers.

As pointed out by various studies \cite{turowski2024generating,proedrou2021comprehensive}, there is a notable shortage of such high-quality data, which hampers the progress in these areas. 
Several factors including technological barriers and privacy concerns lead to the above limitations. Firstly, monitoring sensors and devices, such as smart meters, are often needed for collecting electricity consumption data from households. However, their coverage can be limited in some regions, particularly in developing countries and even within developed economies. For instance, as of 2023, only 54\% of smart meters in the UK operate in smart mode \cite{kerai2023smart}, while Australia reported a smart meter roll-out of approximately 30\% \cite{AEMC2023}. Secondly, the utilization of residential load data is often subject to strict privacy and confidentiality agreements or government regulations due to the sensitive nature of consumer data. Without proper data governance or usage guidelines, even collected load data is often not accessible to stakeholders, including power system operators and researchers.

Significant effort has been dedicated to studying the modeling and generation of residential loads to address the challenges of data scarcity. Existing studies can be broadly categorized into two types based on the methodologies used: bottom-up approaches \cite{dickert2011time,marszal2016household,subbiah2013activity,klemenjak2020synthetic,thorve2023high} and top-down approaches \cite{valverde2012probabilistic,labeeuw2013residential,zufferey2018generating,gu2019gan,liang2022synthesis,claeys2023stochastic,claeys2024capturing,chen2023federated,li2022energy,claeys2023stochastic}. Detailed literature review will be discussed in Section \ref{sec:related_works}. However, existing methods for generating residential loads either require significant data collection and modeling efforts or fail to accurately produce diverse load data. 

To address the research gap, we develop a novel method that leverages a top-down approach using a customized Generative Adversarial Network (GAN)-based framework, namely GAN-based Synthetic Residential Load Pattern (RLP-GAN), to generate household-level load patterns. Our framework integrates over-complete autoencoders and Bidirectional Long Short-Term Memory (Bi-LSTM) networks into the GAN to learn and generate diverse load patterns. These methods, complemented by weakly-supervised adversarial training, enable high accuracy and diversity in the generated patterns, effectively capturing temporal dependencies and lifestyle variety of different households.
Our framework also includes the weight selection technique to avoid mode collapse issues of GAN models. The main contributions of this paper are summarized as follows.
\begin{enumerate}
    \item \textit{Novel Generative Framework for Residential Load Patterns:} The proposed RLP-GAN framework addresses key limitations in existing approaches. Unlike existing methods that require clustering or extensive metadata, our framework integrates over-complete autoencoders with the GAN to directly learn complex load patterns from limited data. This enables more efficient and scalable generation of diverse load profiles.
    \item \textit{Enhanced Data Quality and Diversity:} Our proposed new approach combines weakly supervised training with Fr\`echet distance-based weight selection for load pattern generation. In contrast to existing methods that struggle with mode collapse and pattern quality, our weakly supervised process provides targeted guidance during training, and our weight selection helps identify optimal model states. This enables stable training and better pattern generation, compared to existing methods.
    \item \textit{Comprehensive Validation and Dataset Release:} We validate our approach using real-world data from 417 households, demonstrating significant improvements over state-of-the-art models. Additionally, we release a large-scale synthetic dataset of residential load patterns to support future research in this field.
\end{enumerate}

The evaluation of our proposed RLP-GAN model against four benchmark models yields four key findings regarding the generation of residential load pattern data as follows.
\begin{enumerate}
    \item All evaluated generative models, including our RLP-GAN and four benchmark models (with details in Section~\ref{sec:evaluation}), are able to capture certain load patterns from original data. However, our RLP-GAN outperforms all benchmark models in terms of the ability to generate more diverse load patterns while maintaining a higher level of accuracy in preserving the distributional characteristics of the original load pattern data.
    \item Clustering-based GAN can lead to limitations in generating load patterns that fall between distinct clustering centers. This can affect the overall diversity of the generated load patterns. Our proposed RLP-GAN does not rely on clustering and outperforms benchmark methods.
    \item In comparison to Convolutional Neural Networks (CNN)-based models, LSTM-based GAN models possess superior capabilities in modeling complex load patterns and capturing the distribution of residential load patterns.
    \item The over-complete autoencoder architecture and weakly-supervised training used in our RLP-GAN can effectively capture temporal dependencies presented in load patterns, enhancing the generative performance. 
\end{enumerate}

\section{Related Works} \label{sec:related_works}
This section provides an overview of related works on residential electricity load data generation, and other data generation models applied to broader energy related applications. As summarized in Table \ref{tab:method_comparison}, current methods exhibit a range of trade‑offs in terms of complexity, scalability, and the ability to capture detailed load characteristics. In contrast, the proposed RLP‑GAN framework offers a scalable and non‑intrusive solution that effectively captures temporal dependencies and produces diverse load profiles. The following subsections will provide a detailed discussion of these methodologies, including their respective advantages and limitations.

\begin{table*}[!t]
\centering
{\color{black}               
\caption{Comparisons of Existing Works with the proposed RLP-GAN in this work.}
\label{tab:method_comparison}
\begin{tabular}{p{4cm} p{6.5cm} p{6.5cm}}
\toprule
\textbf{Method} & \textbf{Limitations of Existing Methods} & \textbf{Comparative Advantages of The Proposed RLP-GAN }\\
\midrule
Bottom-up \cite{dickert2011time,marszal2016household,subbiah2013activity,klemenjak2020synthetic,thorve2023high} &
High data collection complexity; intrusive, appliance-level measurements required; scalability issues. &
Scalable; non-intrusive; efficient data utilization.\\[1ex]\midrule

Traditional Machine Learning-based Top-down \cite{valverde2012probabilistic,labeeuw2013residential,zufferey2018generating,claeys2023stochastic} &
Limited detail in short-term dynamics; lower accuracy due to statistical aggregation. &
The ability of capturing explicit, strong temporal dependencies; enhanced accuracy.\\[1ex]\midrule

ACGAN-based \cite{gu2019gan} &
Limited diversity and accuracy due to conditional generation approaches; potential mode collapse risk within clusters. &
No clustering needed; high diversity; robust against mode collapse.\\[1ex]\midrule

DoppelGANger \cite{claeys2024capturing} &
Detailed household metadata required; prone to mode collapse; scalability issues. &
No metadata requirement for generating high load-pattern diversity; stable training and generation; scalable deployment.\\[1ex]\midrule

WGAN-based Conditional Generation \cite{chen2023federated,li2022energy} &
Detailed household metadata required; scalability issues; limited ability to model complex temporal dynamics. &
No metadata requirement for generating high load-pattern diversity; scalable deployment; effective at modeling complex temporal dynamics.\\[1ex]\midrule

RNN-based GAN methods (non-residential load) \cite{asre2022synthetic,yilmaz2023generative,yilmaz2022synthetic} &
Primarily designed for system-level or single-household scenarios, lacking sufficient diversity and detail for residential load data; moderate computational complexity due to recurrent structures. &
Customized for residential load patterns; efficient training, robust temporal modeling, high diversity, and scalability.\\[1ex]\midrule

Diffusion Model \cite{li2024diffcharge} &
Not suitable for tackling the lack of data problem as it requires a large amount of data; slow and computationally intensive sampling process. &
Suitable for scenarios with limited data; faster data generation; computationally efficient.\\[1ex]\midrule

ERGAN \cite{liang2024synthetic} &
Additional computational overhead due to ensemble, inherent risk of mode collapse because of GAN models. & 
Single-model simplicity; no clustering needed; mitigating mode collapse; robust scalability.\\[1ex]
\bottomrule
\end{tabular}
} 
\end{table*}

\subsection{Residential Load data Generation Methodology}
Residential electricity load data modeling and generation techniques can be broadly categorized into two main approaches: \textit{top-down} and \textit{bottom-up} approaches. 

\subsubsection{Bottom-up Approach}
This approach, used in \cite{dickert2011time,marszal2016household,subbiah2013activity,klemenjak2020synthetic,thorve2023high}, typically models the electricity load for individual appliances or users, treating each as an independent entity and then aggregating their energy consumption to estimate the total household load. Some variations of this bottom-up approach have been developed to model residential load, focusing on appliance-level power consumption, usage patterns, and operating times. These approaches aim to capture realistic and granular load profiles by considering detailed characteristics of individual appliances and activities.

While bottom-up methods have demonstrated their efficacy in generating synthetic data with high realism, diversity, and similarity, the process of data collection and modeling can often be expensive and time-consuming. Many methods that utilize the bottom-up approach depend on additional intrusive sensors for individual data collection, which are then aggregated to generate household-level load data. Even non-intrusive methods, such as \cite{klemenjak2020synthetic},  do not address issues related to data collection and load modeling. As a consequence, methods that employ the bottom-up approach often prove difficult to adapt to a large scale of households. As such, our proposed framework is specifically designed to overcome these challenges associated with residential load data collection.

\subsubsection{Top-down Approach}
In contrast, top-down approaches treat the electricity load of each household as a unified entity, without considering individual appliance-level consumption during load generation. Valverde et al. in \cite{valverde2012probabilistic} used Gaussian Mixture Models (GMM) to model household electricity load. Labeeuw et al. in \cite{labeeuw2013residential} introduced a pre-clustering method that groups load data using the expectation-maximization clustering algorithm, followed by two Markov models to simulate the electricity consumption of each cluster. Zufferey et al. in \cite{zufferey2018generating} employed adaptive Markov chain models to generate realistic load profiles for non-metered households. Claeys et al. in \cite{claeys2023stochastic} introduced a wavelet-based decomposition method to capture short-term temporal dynamics more effectively than conventional clustering, which often smooths out variability. By separating daily profiles into low-frequency and high-frequency components, the wavelet method preserves intraday fluctuations. However, combining low-frequency and high-frequency components from different households can still result in artificial peaks or mismatches in short-term behaviors. Although rescaling and shifting are applied to align peak demands, these adjustments do not fully capture the exact timing of short-term variations.

The top-down approach is often seen as more cost-effective for both data collection and modeling, as it only requires aggregated load data. However, traditional machine learning methods frequently face issues, such as limited accuracy and a loss of detailed load information. To address these challenges, Gu et al. in \cite{gu2019gan} proposed using GANs along with a pre-clustering approach. Although innovative, this method is limited by the inherent diversity of residential load data, which arises from the wide range of housing characteristics and variations in user lifestyles. This diversity reduces the method's accuracy and ability to generate diverse load profiles, especially when significant differences exist in consumption patterns across clusters. Claeys et al. in \cite{claeys2024capturing} improved on this using a more advanced GAN architecture, specifically the DoppelGANger model, designed to capture complex temporal dynamics. This work builds upon earlier studies \cite{chen2023federated,li2022energy} that generated load profiles based on metadata, such as appliance ownership and household characteristics, to generate more diverse and realistic profiles. However, collecting this metadata is generally difficult, limiting the model's applicability. To overcome these limitations, a recent study by Liang et al. \cite{liang2024synthetic} extended the pre-clustering paradigm through an ensemble of recurrent GANs trained on clustered load data. This approach can enhance pattern diversity and eliminate metadata requirements. But the pre-clustering approach itself may introduce scalability constraints and biases stemming from rigid cluster boundaries—a potential limitation for heterogeneous or dynamically evolving load profiles. Furthermore, the GAN-based framework inherits mode collapse risks, underscoring the need for enhanced robust diversity-preserving mechanisms.

\subsection{Energy Data Research Using Generative Models}

Advanced generative models, such as GANs and diffusion models, have significantly impacted data generation methodologies across various research domains. 
Originally showcasing their effectiveness in healthcare data protection \cite{amrit2023embedr}, \textcolor{black}{consumer electronics waste management \cite{10552419}}, intelligent transportation systems \cite{li2024gan}, and wireless communication \cite{njima2022dnn}, these techniques have also been adapted for energy-related applications.
In energy systems, GANs have been pivotal in addressing the challenges of data generation, particularly for load data, renewable generation data, and electric vehicle (EV) charging data.

For instance, Asre et al. in \cite{asre2022synthetic} proposed a more advanced time-variant GAN for generating state-level load data, specifically related to electricity consumption across various Australian states. This approach shares similarities with \cite{yilmaz2023generative} on Turkey's electricity market, as both studies focus on system-level load generation using GANs. However, it is important to note that the system-level load data mentioned in this context represents the aggregated electricity consumption across diverse categories, including residential, commercial, industrial, and other sectors. Consequently, load data at this level tend to exhibit high stability and relatively limited patterns and demand variations. In another line of research, \cite{yilmaz2022synthetic} conducted a benchmarking study to evaluate the performance of different GAN variants in generating single-household load data. However, it was highlighted by \cite{wang2021identifying} that households with distinct socioeconomic characteristics can exhibit unique load patterns. In both of the aforementioned scenarios, the data derived from individual households lack the necessary diversity to fully demonstrate the method's ability to encompass the intricate variability of load patterns across residential households.

Alongside GANs, diffusion models, which are being extensively employed in the computer vision community, have gained attention for their effectiveness in generating high-fidelity data. These models, known for their precision in image generation, gradually reverse a diffusion process to transform random noise into structured data, achieving impressive results in applications like simulating EV charging sessions. Li et al. in \cite{li2024diffcharge} showcased the potential of diffusion models in creating realistic EV charging data, which inherently involves complex temporal dynamics and variability. Recent studies, such as \cite{akbar2023beware}, have shown that a drawback of current diffusion models is their need for relatively large datasets to achieve effective training outcomes. This characteristic limits their applicability in scenarios like residential load data generation, where the challenges to tackle are the impracticality of collecting extensive datasets. Moreover, the slow sampling process of diffusion models \cite{yang2023diffusion}, though secondary, further complicates their use in applications where rapid data generation is crucial for real-time simulation and analysis.

Given these considerations, our research focuses on leveraging the strengths of GANs to effectively overcome data scarcity challenges and capture complex temporal dynamics in residential electricity usage. By addressing these issues, we aim to develop a framework for more accurate and efficient synthetic load pattern generation. We will detail the framework's design and implementation in the following sections.

\section{Methodology of Residential Load Pattern Generation}
To overcome the data scarcity of residential load data, we develop RLP-GAN, which incorporates overcomplete autoencoders within the GAN architecture and a weakly-supervised learning approach during the training process to achieve finer control over the network dynamics.

\begin{figure*}[!t]
\centering
\includegraphics[width=0.8\textwidth]{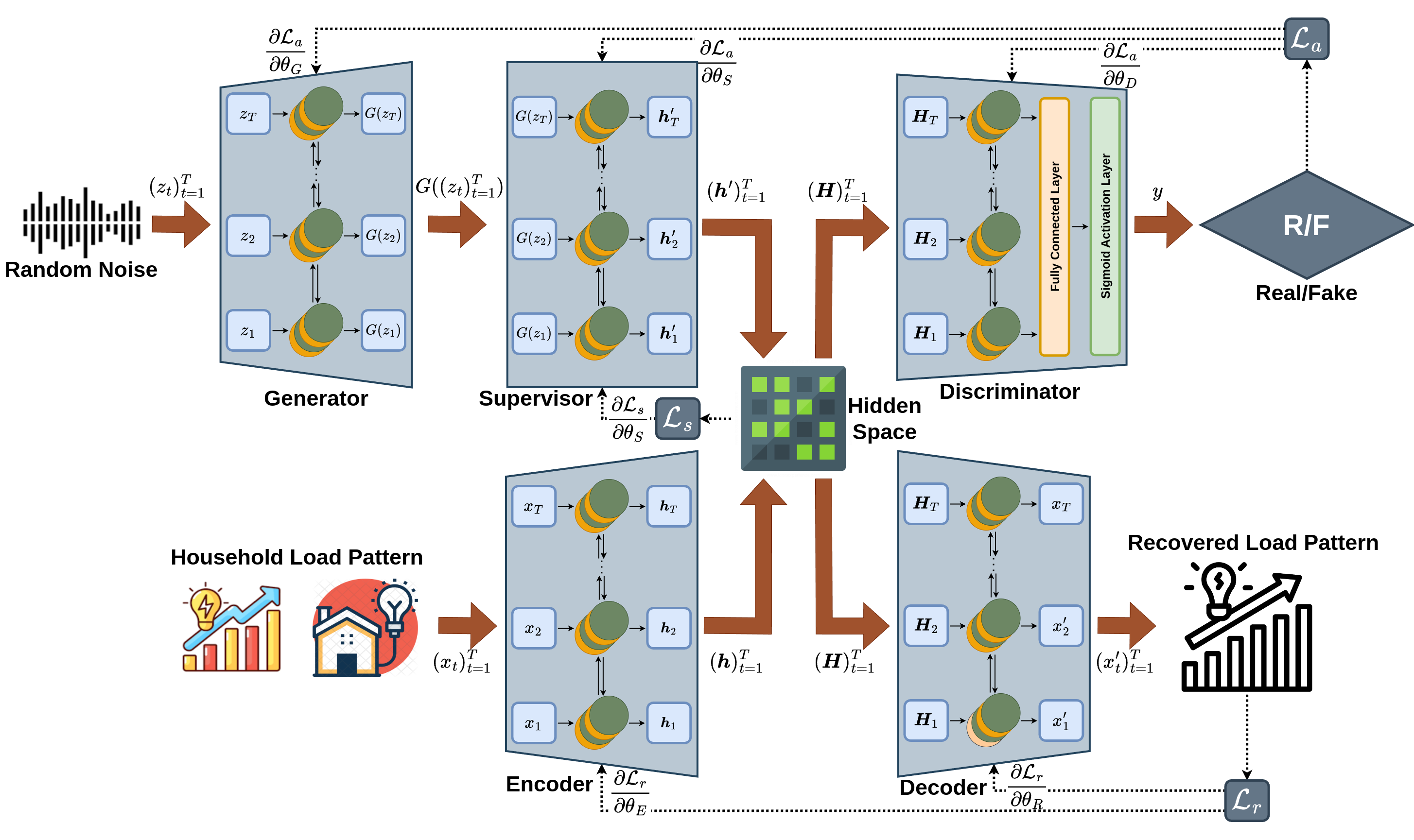}
\caption{Overview of the RLP-GAN framework. The generator creates synthetic load patterns from noise vectors, which are refined by the supervisor. The encoder processes real data into hidden representations. The discriminator evaluates these representations, and the decoder reconstructs the load patterns.}
\label{ELP-GAN_Architecture}
\end{figure*}

The architecture of the RLP-GAN model is illustrated in Figure \ref{ELP-GAN_Architecture}. In the model, the original household load pattern will go through an encoding process to construct a hidden representation while the random noise can go through generating process to generate a hidden representation as well. These hidden representations can be evaluated by the discriminator to determine whether they are constructed from a real or generated load pattern. The decoding process can convert the hidden representation back into the load pattern format. The training process of the model, including the loss calculation and the model weight updates, are also illustrated in Figure \ref{ELP-GAN_Architecture}. In contrast to \cite{chen2018model,gu2019gan} using  CNN to perform energy data generation, where temporal information can only be extracted from local neighborhoods, we adopt Bi-LSTM network to better capture dependencies across the complete sequence. Detailed explanation of the RLP-GAN model in Figure~\ref{ELP-GAN_Architecture} will be presented in Sections~\ref{subsec:TwoStageGeneration} and \ref{subsec:Autoencoders_With_GAN}.

\subsection{Problem Formulation}
Given the time-dependent nature of the load pattern, we formulate the load pattern generation problem into a time-series generation problem.
For each load pattern sequence as a time series sequence with length $T$, denoted as $(x_t)_{t=1}^T$. 

In load generation research, the generated data is expected not only to accurately reflect the diverse load patterns across various households but also to retain the temporal dynamics within the sequence of load patterns. Therefore, we develop the residential load pattern generation method to achieve the following two goals.

Goal \#1 aims to learn a probability density of temporal features of residential load patterns from generated dataset (across households), denoted as $p'_x((x_t)_{t=1}^T)$, to best approximate the probability density in the training dataset, denoted as $p_x((x_t)_{t=1}^T)$, such that similarity is achieved at the dataset level. Goal 1 can be formulated as the following objective:
\begin{equation} \label{eq:goal1}
\min_{p_x'}~\text{Dist}(p_x((x_t)_{t=1}^T)||p_x'((x_t)_{t=1}^T)),
\end{equation}
which minimizes the distance between $p_x((x_t)_{t=1}^T)$ and $p'_x((x_t)_{t=1}^T)$ by optimizing $p_x'$.

Goal \#2 aims to capture temporal dynamics at the sequence level of load patterns. Specifically, we focus on conditionals and learn a conditional probability density $p'(x_{t=H}|(x_t)_{t=1}^{H-1})$ to best approximate the conditional probability of load pattern sequences in the training dataset $p(x_{t=H}|(x_t)_{t=1}^{H-1})$ at timesteps $H = 2,...,T$, such that the sequence-level similarity can be achieved. Goal 2 can be formulated as the following objective:
\begin{equation} \label{eq:goal2}
\min_{p_x'}~\text{Dist}(p_x(x_{t=H}|(x_t)_{t=1}^{H-1})||p_x'(x_{t=H}|(x_t)_{t=1}^{H-1})),
\end{equation}
which minimizes the distance between $p(x_{t=H}|(x_t)_{t=1}^{H-1})$ and $p'(x_{t=H}|(x_t)_{t=1}^{H-1})$ by optimizing $p_x'$.

\subsection{Two-Stage Generation Process}
\label{subsec:TwoStageGeneration}
As GAN has been showing powerful capability in learning the underlying distribution of given data, we choose GAN as the base for building the synthetic residential load pattern generation model.
The base of our model also consists of the two main components in the standard GAN: the generator denoted as a differentiable function $G$ with parameters $\theta_G$ and and the discriminator denoted as a differentiable function $D$ with parameters $\theta_D$, as shown in Figure \ref{ELP-GAN_Architecture}. The generator and discriminator are trained against each other, where the goal of the Generator is to generate synthetic data that is real enough to fool Discriminator, and the goal of the Discriminator is to distinguish between real and generated data.

In addition to the standard GAN, we introduce weakly-supervised learning as a supervisor unit incorporated with the generator unit to construct a two-stage data generation process, inspired by TimeGAN \cite{yoon2019time}. As shown in Figure \ref{ELP-GAN_Architecture}, we denote the supervisor as a differentiable function $S$ with parameters $\theta_S$. The adversarial training process is also changed, as the generator and supervisor together form a two-stage generation process and then pass generated results to the discriminator for classification. The supervisor is responsible for finely adjusting the temporal dynamics within the sequence generated by the generator, so that the sequence-level temporal dependencies can be well retained during the generation process. 

\subsection{Over-Complete Autoencoders With GAN} \label{subsec:Autoencoders_With_GAN}

To further improve our model's capability of capturing diverse load patterns across a large number of households, we incorporate over-complete autoencoders in our GAN model. Previous approaches typically rely on direct generation in the data space, such as in \cite{gu2019gan, claeys2024capturing}, but often struggle to capture the full dynamics of residential load patterns. While adopting under-complete autoencoders with GAN has demonstrated benefits in various applications \cite{larsen2016autoencoding,dumoulin2016adversarially}, our work uses over-complete representations to enhance residential load pattern generation.
Specifically, as shown in Figure \ref{ELP-GAN_Architecture}, the over-complete autoencoder in our RLP-GAN model includes an encoder denoted as a differentiable function $E$ with parameters $\theta_E$ and a decoder denoted as a differentiable function $R$ with parameters $\theta_R$. In our work, we integrate the over-complete autoencoders to conduct the sparse hidden representation $h_t$ of the single temporal feature (i.e., electricity consumption) from the original sample $x_t$ in the input space. The over-complete autoencoders construct the representation in the hidden space where it has higher dimensions compared to the original sample in the input space. In contrast to the commonly used under-complete autoencoders, the key advantage of over-complete autoencoders lies in their ability to construct sparse representations \cite{bazzi2016sparse} that can better model complex and diverse structures inherent in input load patterns. Such a feature-rich representation is well-suited for the top-down approach used in the residential load pattern generation.

\subsection{Three-Stage Weakly Supervised Adversarial Training}
Weak supervision in our framework is a hybrid training paradigm that combines partial temporal guidance with unsupervised adversarial learning. Unlike conventional methods that incorporate unsupervised feature extraction processes to address the scarcity of labeled data, our approach uses the encoded temporal dependencies within sequences to implicitly guide the generator. Specifically, the supervisor network predicts the hidden representation in the next timestep $h_{t=H}$ using only the preceding sequence $(h_t)_{t=1}^{H-1}$. This provides indirect supervision by enforcing stepwise consistency between real and synthetic sequences, without relying on explicit labels. The weakly supervised training methodology in our proposed RLP-GAN consists of three steps: autoencoder training, supervisor training, and joint training.

\subsubsection{Step 1: Autoencoder Training}
The autoencoder is trained during the first stage of the training process. It is responsible for creating a sequence of vectors $(\boldsymbol{h}_t)_{t=1}^T$ as the hidden representation based on the original load pattern data $(x_t)_{t=1}^T$, which is formulated as
\begin{equation} \label{eq:encoding}
(\boldsymbol{h}_t)_{t=1}^T = E((x_t)_{t=1}^T; \theta_E).
\end{equation}

Then the decoder reconstructs the load pattern data from its hidden representation $(h_t)_{t=1}^T$ to $(x'_t)_{t=1}^T$, which is formulated as
\begin{equation} \label{eq:decoding}
(x'_t)_{t=1}^T= R((\boldsymbol{h}_t)_{t=1}^T; \theta_R) = R(E((x_t)_{t=1}^T; \theta_E); \theta_R).
\end{equation}

The training process should allow the encoder to extract meaningful information, while the decoder can accurately reconstruct the original data from the hidden representation. Therefore, the autoencoder training aims to minimize the reconstruction loss function $\mathcal{L}_r$ defined as
\begin{equation} \label{eq:reconstruction_loss}
\mathcal{L}_r = \mathbb{E}_{(x_t)_{t=1}^T \sim p_x}\left[\sum_{t=1}^T||x_t - x_t'||^2\right],
\end{equation}
which calculates the distance between original input data and the reconstructed data. 

As shown in Figure \ref{ELP-GAN_Architecture}, the training objective is to minimize $\mathcal{L}_r$ by optimizing $\theta_E$ and $\theta_R$ in the autoencoder.

\subsubsection{Step 2: Supervisor Training}
Incorporating weak supervision in GANs can enhance the ability to capture the stepwise conditional distributions in the data. In this step, supervisor training is performed in a supervised manner to guide the generated sequence to capture stepwise temporal correlation before proceeding to the unsupervised adversarial training process. During the supervised training process, the supervisor $S$ takes information from part of the encoded data $(\boldsymbol{h}_t)_{t=1}^{H-1}$ (instead of from the generator) as its input to generate ${\boldsymbol{h}'}_{t=H}^{\text{sup}}$ in the hidden space at next timestep $H \in \{2...T\}$, written as
\begin{equation} \label{eq:next_step_est}
{\boldsymbol{h}'}_{t=H}^{\text{sup}}=S((\boldsymbol{h}_t)_{t=1}^{H-1}; \theta_S).
\end{equation}

The generated vector ${\boldsymbol{h}'}_{t=H}^{\text{sup}}$ in the hidden space is expected to be close to the hidden representation $\boldsymbol{h}_{t=H}$ of the original data, given the input $(\boldsymbol{h}_t)_{t=1}^{H-1}$. In other words, the difference between the generated next-stage vector and the encoded next-stage vector should be minimized. The objective of the supervisor training is to minimize the difference at the sequence level, such that we can formulate the supervisor loss  function $\mathcal{L}_s$ to minimize the difference between time-series sequences $(\boldsymbol{h}_t)_{t=2}^{T}$ and $({\boldsymbol{h}'}_t^{\text{sup}})_{t=1}^{T-1}$, i.e.,
\begin{equation} \label{eq:supervised_loss}
\mathcal{L}_s = \mathbb{E}_{(x_t)_{t=1}^T \sim p_x}\left[\sum_{t=1}^{T-1}||\boldsymbol{h}_{t+1} - {\boldsymbol{h}'}_t^{\text{sup}}||^2\right].
\end{equation}

As shown in Figure \ref{ELP-GAN_Architecture}, the supervisor training aims at minimizing $\mathcal{L}_s$ by optimizing the supervisor weights $\theta_S$.

\subsubsection{Step 3: Joint Training}
Joint Training is designed to facilitate a weakly-supervised iterative process that combines supervisor training and unsupervised adversarial training techniques.
During the unsupervised adversarial training, the two-stage generation process formed with the generator and supervisor is trained against the discriminator in the hidden space. The generator begins by taking a random noise sequence represented as $(z_t)_{t=1}^{T}$ under a known distribution $p_z$ denoted as $(z_t)_{t=1}^{T} \sim p_z$. The generator then produces sequences, which are subsequently refined by the supervisor. The resulting data representation $({\boldsymbol{h}'}_t)_{t=1}^T$ is formulated as 
\begin{equation} \label{eq:generating}
({\boldsymbol{h}'}_t)_{t=1}^T=S(G((z_t)_{t=1}^T;\theta_G);\theta_S).
\end{equation}

Discriminator $D$ takes hidden representation $(\boldsymbol{h}_t)_{t=1}^T$ from real sample $(x_t)_{t=1}^T$ or generated representation $({\boldsymbol{h}'}_t)_{t=1}^T$ as the input, and calculates the probability of whether the inputted representation is from real data. This probability can be formulated as
\begin{equation}
y = D((\boldsymbol{H}_t)_{t=1}^T;\theta_D),
\end{equation}
where $(\boldsymbol{H}_t)_{t=1}^T$ represents either $(\boldsymbol{h}_t)_{t=1}^T$ or $(\boldsymbol{h}'_t)_{t=1}^T$.

During adversarial training, the discriminator is expected to maximize the value of $D((\boldsymbol{h}_t)_{t=1}^T;\theta_D)$ and minimize the value of $D(({\boldsymbol{h}'}_{t})_{t=1}^T;\theta_D)$. Hence, the loss function of discriminator, denoted as $\mathcal{L}_D$, 
can be formulated as
\begin{equation} \label{eq:discriminator_loss}
\begin{aligned}
\mathcal{L}_D = -\mathbb{E}_{(x_t)_{t=1}^T \sim p_x}[-\log D((\boldsymbol{h}_t)_{t=1}^T;\theta_D)] \\ -\mathbb{E}_{(z_t)_{t=1}^T \sim p_z}[\log (1-D(({\boldsymbol{h}'}_{t})_{t=1}^T;\theta_D))].
\end{aligned}
\end{equation}

At the same time, the generator and supervisor are expected to maximize the value of $D(({\boldsymbol{h}'}_t)_{t=1}^T;\theta_D)$. Hence, we formulate the loss function of the generation process $\mathcal{L}_G$ as 
\begin{equation} \label{eq:generator_loss}
\begin{aligned}
\mathcal{L}_G = \mathbb{E}_{(z_t)_{t=1}^T \sim p_z}[\log(1-D(({\boldsymbol{h}'}_{t})_{t=1}^T;\theta_D))].
\end{aligned}
\end{equation}

The adversarial training involves the discriminator $D$ and the generation process $G$ with $S$ and can be considered as a two-player minmax game. The loss function associated can be expressed as a minmax value function $\mathcal{L}_a$ formulated as
\begin{equation} \label{eq:adversarial_loss}
\begin{aligned}
\mathcal{L}_a = \min _{G} \max _{D} V(D, G) = \mathbb{E}_{(x_t)_{t=1}^T \sim p_x}[\log D((\boldsymbol{h}_t)_{t=1}^T;\theta_D)] \\
+ \mathbb{E}_{(z_t)_{t=1}^T \sim p_z}[\log (1-D(({\boldsymbol{h}'}_t)_{t=1}^T;\theta_D))].
\end{aligned}
\end{equation}

The goal of adversarial training is to achieve Nash equilibrium, where the generation process is capable of producing synthetic data that closely resemble real data, and cannot be distinguished by the discriminator. However, relying solely on the feedback provided by the discriminator is insufficient for capturing the stepwise conditional distributions in the data. To address this limitation, in addition to the unsupervised adversarial training utilizing $\mathcal{L}_a$, a supervised training process is conducted using the supervised loss function $\mathcal{L}_s$. This additional training process facilitates the updating of the supervisor weight $\theta_S$ along with the adversarial training. Combining supervised and unsupervised training methods serves to ensure that the generated data exhibit similar stepwise transitions as those observed in the original data.

\subsection{Model Weight Selection}
The GAN training process is notoriously unstable and prone to mode collapse, where the generator produces similar outputs regardless of the input, especially when overtrained \cite{arjovsky2017wasserstein}. Existing approaches typically address this through architectural modifications or loss function engineering. In contrast, we introduce a novel weight selection approach based on distribution similarity. Our approach differs from conventional methods by providing a more direct means to monitor and maintain generation quality throughout the training process. In our task of generating close to real synthetic load pattern data, it is essential that the model learns the distribution of real electricity load pattern data implicitly. However, the quality of the generated data cannot be adequately reflected using $\mathcal{L}_D$ alone. To address this limitation, we draw inspiration from the widely-used evaluation method in image data generation \cite{kim2021exploiting,zhang2021defect} and the ``Fr\'echet Inception Distance (FID)" \cite{heusel2017gans}. We design a model weight selection method by selecting the model weight with the lowest measured distance between distributions of real and generated data. As the study in \cite{munkhammar2014characterizing} showed that the aggregate distribution of multiple uncorrelated households approaches a Gaussian, we assume that the load pattern data follows a multivariate Gaussian distribution, represented by the mean and covariance polynomials for practical purposes. We calculate the Fr\'echet distance using the formula proposed in \cite{dowson1982frechet} as 
\begin{equation} \label{eq:Frechet_distance}
\begin{array}{l}
\mathcal{D}_{\text{Fr\'echet}}^2 = ||\mu_{X_{\text{real}}} - \mu_{X_{\text{fake}}}||^2 \\
+ tr\left(\sum_{X_{\text{real}}} + \sum_{X_{\text{fake}}} - 2\left(\sum_{X_{\text{real}}}\sum_{X_{\text{fake}}}\right)^{\frac{1}{2}}\right),
\end{array}
\end{equation}
in which the original samples are represented as $X_{\text{real}}$ while the generated samples are represented as $X_{\text{fake}}$. The Fr\'echet distance $\mathcal{D}_{\text{Fr\'echet}}^2$ calculates the distance between two Multivariate Gaussian Distribution with means $\mu_{X_{\text{real}}}$ and $\mu_{X_{\text{fake}}}$, covariance matrices $\sum_{X_{\text{real}}}$ and $\sum_{X_{\text{fake}}}$.

In our experiments, we monitor the change in the value of $\mathcal{D}_{\text{Fr\'echet}}^2$ and select the model with the lowest $\mathcal{D}_{\text{Fr\'echet}}^2$ as our best-performing model weight. As we will present in Section~\ref{subsec:modelweight}, our results demonstrate that evaluating the quality of data generated using Fr\'echet distance $\mathcal{D}_{\text{Fr\'echet}}^2$ is a more informative way than relying solely on $\mathcal{L}_D$, and it correlates well with the performance measurements.

\section{Experiment Setup and Evaluation}\label{sec:evaluation}

\subsection{Data Description and Preparation}

We use real-world data from the Pecan Street dataset \cite{WinNT}  
to train and evaluate the RLP-GAN framework. The dataset contains detailed smart meter data, capturing hourly electricity consumption (in kWh) from 417 households over the year 2017. To ensure data quality, we removed any time-series data with missing points and restructured the remaining data into daily load patterns. As part of the preprocessing, we applied a Min-Max scaler to normalize each daily load pattern to a range of 0 to 1. After this cleaning and scaling process, we obtained a total of 150,727 household days of high-quality data. This processed dataset provides a solid foundation for training and evaluating the RLP-GAN model, enabling the generation of realistic and diverse synthetic residential load data that reflect actual consumption patterns.

\subsection{Benchmark Models}

To evaluate our methodology, we compare it with four generative models recently applied in energy applications, representing state-of-the-art approaches to synthetic data generation.
The Auxiliary Classifier Generative Adversarial Network (ACGAN) \cite{odena2017conditional}, a CNN-based conditional GAN, is utilized for its proficiency in handling pre-clustered class labels. The Wasserstein Generative Adversarial Network (WGAN) \cite{chen2018model}, known for using the Wasserstein distance to enhance the stability of GAN training, is chosen to represent advanced CNN-based GANs in our benchmarks. The Continuous RNN-GAN (C-RNN-GAN) \cite{mogren2016c}, which incorporates LSTM units, is selected for its ability to effectively capture temporal features. We also incorporate a Denoising Diffusion Probabilistic Model (DDPM) \cite{ho2020denoising}, a CNN-based diffusion model that has recently shown state-of-the-art performance in generating high-quality image data through iterative noise reduction. These models serve as benchmarks to validate the effectiveness of our proposed solution under controlled conditions.

\subsection{Evaluation Criteria and Methodology}
To assess our generator’s performance, we introduce a comprehensive evaluation framework focusing on two key criteria: diversity and similarity.
To this end, we expect the generated data samples to be diverse enough to capture different electricity load patterns from various households, while at the same time, the generated dataset should follow similar statistical properties as the real data to reflect the distribution of the original dataset. Hence, we developed our evaluation methods which can address the two criteria at two levels following two approaches: 
\begin{itemize}
    \item \textbf{Sequence Level Evaluation:} We compare the similarity and diversity of the generated data with the original data at the sequence level.
    \item \textbf{Aggregated Level Evaluation:} We aggregate all sequences in the dataset to perform the evaluation on the aggregated level.
\end{itemize}

\subsubsection{Sequence Level Evaluation} \label{subsubsec:Sequence_Level_Evaluation}
By evaluating the generated data at the sequence level, we are able to compare the generated sequences with the original load pattern sequences. In an ideal scenario, each generated sequence should closely match at least one real sequence. To achieve this objective, we have developed two methods as described below.

\begin{itemize}
    \item 
    \textbf{Visual comparison of real and fake sample sets}.     
    We commence our comparison by applying a K-means clustering algorithm (with $K$ set to 4) to the real samples from the testing dataset. From each cluster, a sample is randomly chosen and the corresponding synthetic sample is identified based on the principle of minimum Euclidean distance, ensuring that each real sample is paired with its closest generated counterpart. A visual assessment is conducted to evaluate pattern similarity.
    Additionally, we use auto-correlation to compare the correlations between observations at different points in the time series. The auto-correlation 
    of load profile $(x_i)_{i=1}^{t}$ with mean $\mu_x$ and standard deviation $\sigma_x$ can be calculated as
    \begin{equation} \label{eq:autocorr}
        \text{Autocorr}(k) = \frac{1}{t-k}\sum_{i=1}^{t-k}\frac{(x_i-\mu_x)(x_{i+k}-\mu_x)}{\sigma_x^2}.
    \end{equation}

    \item \textbf{High Dimensional Load Pattern Data Visualization}. 
    We use dimension reduction techniques, including both linear (PCA) and non-linear (T-SNE) \cite{van2008visualizing} methods, to visualize and compare the similarity and diversity between the original and generated data samples using scatter plots. This approach increases the robustness of our evaluation method by utilizing multiple techniques to provide a comprehensive assessment of our generative model's performance.
\end{itemize}

\subsubsection{Aggregated Level Evaluation} \label{subsubsec:Aggregated_Evaluation}
The evaluation of generated data at the aggregated level facilitates the measurement of statistical resemblance between the generated dataset and the real test dataset by consolidating all load pattern sequences. To achieve this, we develop two types of measurements for aggregated evaluation.
\begin{itemize}
    \item \textbf{Statistical Resemblance Measurement}. 
    To assess the aggregated profiles of both the original and generated data samples, we calculate their key statistical properties including mean, standard deviation, minimum and maximum values, as well as the first, second, and third quartiles. To further substantiate our evaluation of their similarity, we undertake a comparative analysis of the probability distribution for each value. This comparison is visually represented via the Cumulative Distribution Function (CDF) graph, facilitating a visual assessment of their resemblances.
    \item \textbf{Distance Measurement}. 
    To supplement our evaluation with numerical measures, we quantify the differences between the original dataset distribution ($P(x)$), and the generated dataset distribution, ($Q(x)$) assumed in the same probability space $X$, using several metrics calculated on the aggregated datasets.

    First, we calculate the Root Mean Square Error (RMSE) between the mean profiles, denote as $\mu_{X_{test}}=(\mu_{X_{test}}^i)_{i=1}^t$ and $\mu_{X_{gen}}=(\mu_{X_{gen}}^i)_{i=1}^t$ of of the testing dataset $X_{test}$ and generated dataset $X_{gen}$ respectively:
    \begin{equation} \label{eq:RMSE}
        \text{RMSE}(\mu_{X_{test}}, \mu_{X_{gen}}) = \sqrt{\frac{1}{t}\sum_{i-1}^t(X_{test}^i-X_{gen}^i)^2}.
    \end{equation}

     Second, to gain asymmetric insights into how the distributions differ, we compute the Kullback-Leibler (KL) divergence. This divergence measures the difference between two probability distributions. We report it in both directions:
     \begin{align} 
        D_{\text{KL}}(P || Q) = \sum_{x \in {X}} P(x) \log \frac{P(x)}{Q(x)},\label{eq:fwd_kl} \\
        D_{\text{KL}}(Q || P) = \sum_{x \in {X}} Q(x) \log \frac{Q(x)}{P(x)},\label{eq:rev_kl}
    \end{align}
    where the Forward KL is represented as $D_{\text{KL}}(P || Q)$ and the Reverse KL is represented as $D_{\text{KL}}(Q || P)$.

    In addition to the KL divergence, we compute the Jensen-Shannon (J-S) distance formulated as
    \begin{equation}
        d_\text{JS}=\sqrt{\text{JS}(P(x)||Q(x))},
    \end{equation}
    which provides a symmetric measure of similarity between two distributions and is defined as the square root of the J-S Divergence (JS):
    \begin{equation} \label{J-S}
    \begin{aligned}
    \text{JS}(P(x)||Q(x)) = \frac{1}{2}\sum_{x \in X}\left[P(x)\log\frac{2P(x)}{P(x)+Q(x)} + \right. \\ \left.Q(x)\log\frac{2Q(x)}{P(x)+Q(x)}\right].
    \end{aligned}
    \end{equation}
\end{itemize}

\subsection{Model Configuration}

Table \ref{table:rlp-gan_structure} provides an overview of the RLP-GAN model architecture, detailing the configuration of key components: the generator, discriminator, encoder, decoder, and supervisor. As training hyperparameters is essential to optimize model performance and stability, Table \ref{table:rlp-gan_hyperparameters} summarizes the chosen training parameters, including the batch size, learning rate, 
 and optimizer. 
The parameters in Table \ref{table:rlp-gan_structure} and \ref{table:rlp-gan_hyperparameters} are identified using a focused grid search strategy. Recognizing that exhaustive searches are often impractical for complex GANs, our approach involves firstly identifying critical parameters (including hidden dimensions, learning rate, and batch size) and determining constrained working ranges for them based on preliminary experiments observing factors, such as training stability, convergence speed, and model capacity. The grid search then systematically evaluates combinations, primarily focusing on learning rate and batch size, within these narrowed, empirically justified ranges to find settings that balanced convergence, stability, and output quality.
These settings are determined through experiments to balance convergence and output quality, notably accounting for dataset characteristics, such as size and complexity. For instance, smaller datasets may require lower model capacity (e.g., fewer hidden units) to avoid overfitting. In contrast, larger or more geographically diverse datasets can benefit from increased model complexity to capture nuanced temporal dependencies.

\begin{table}[!t]
\centering
\caption{RLP-GAN Model Structure.}
\begin{tabular}{|c|c|c|}
\hline
\textbf{Component}         & \textbf{Parameter}           & \textbf{Value} \\ \hline
\multirow{2}{*}{Generator} & Bi-LSTM Layer             & 32 hidden units, 5 layers                    \\ 
                           & Activation Function          & Sigmoid                   \\ \hline
\multirow{2}{*}{Discriminator} & Bi-LSTM Layer               & 32 hidden units, 5 layers \\ 
                           & Activation Function          & Sigmoid              \\ \hline
\multirow{2}{*}{Encoder}   & Bi-LSTM Layer             & 32 hidden units, 5 layers                     \\ 
                           & Activation Function          & Sigmoid                   \\ \hline
\multirow{2}{*}{Decoder}   & Bi-LSTM Layer                 & 32 hidden units, 5 layers \\ 
                           & Activation Function          & Sigmoid                \\ \hline
\multirow{2}{*}{Supervisor} & Bi-LSTM Layer                & 32 hidden units, 5 layers                    \\ 
                           & Activation Function          & Sigmoid                   \\ \hline
\end{tabular}
\label{table:rlp-gan_structure}
\end{table}

\begin{table}[!t]
\centering
\caption{RLP-GAN Training Hyperparameters.}
\begin{tabular}{|l|c|}
\hline
\textbf{Hyperparameter}   & \textbf{Value} \\ \hline
Batch Size                & 128            \\ 
Autoencoder Training Epochs                    & 1000          \\ 
Supervisor Training Epochs                    & 1000          \\ 
Adversarial Training Epochs                    & 10000          \\ 
Learning Rate             & 0.0005          \\ 
Optimizer                 & Adam           \\ 
Loss Function             & Binary Cross-Entropy \\ 
Noise Type                & Gaussian       \\ \hline
\end{tabular}
\label{table:rlp-gan_hyperparameters}
\end{table}

\section{Results And Analysis}

\subsection{Effectiveness Proof of Model Weight Selection Methods}\label{subsec:modelweight}
\begin{figure}[!t]
\centering
\includegraphics[width=\columnwidth]{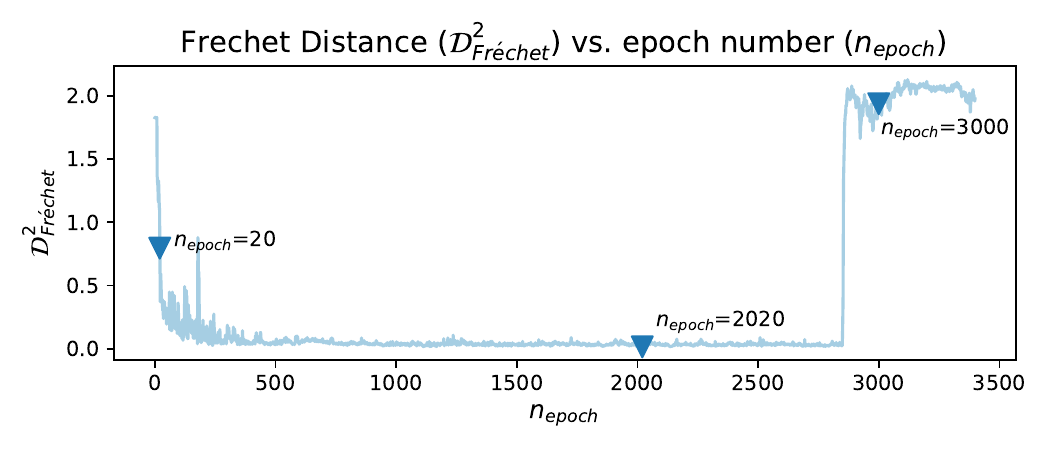} 
\includegraphics[width=\columnwidth]{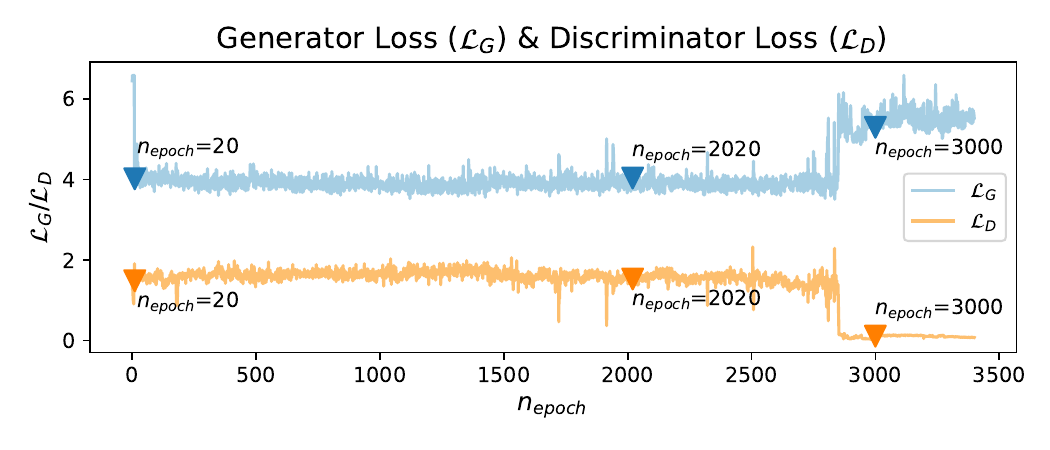}
\includegraphics[width=\columnwidth]{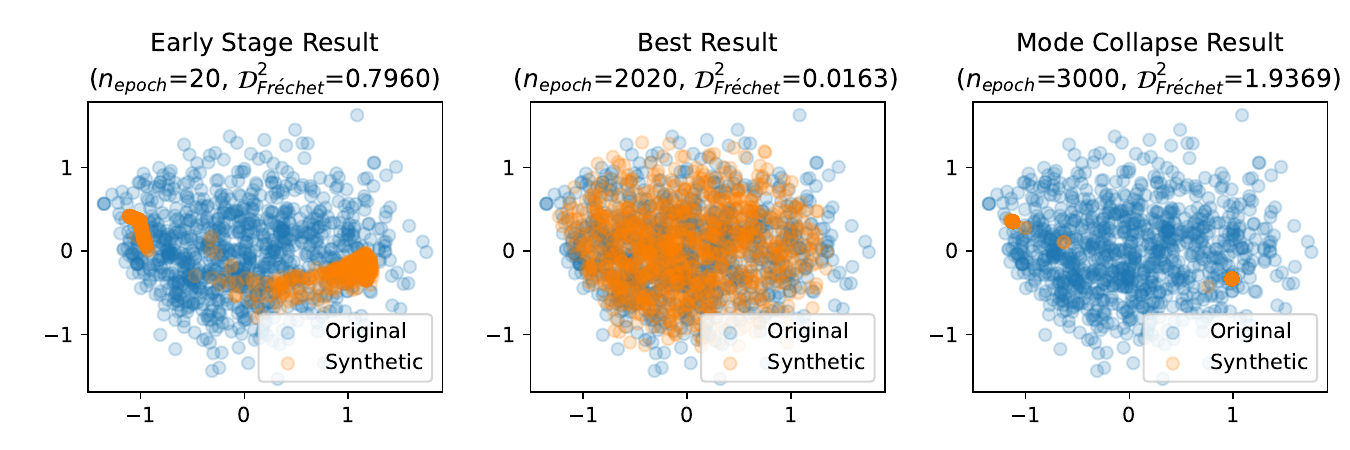}
\caption{Fr\'echet distance and loss value change along with dimension reduced result Visualization during RLP-GAN training. The epoch number is denoted as $n_{epoch}$.}
\label{model_selection}
\end{figure}

To show the effectiveness of our model selection method, we demonstrate a training example of RLP-GAN with the mode collapse problem as shown in Figure \ref{model_selection}. We observe that at the early stage of the training process, the Generator and Discriminator losses are fluctuating, where it is hard to distinguish the learning progress of the generator by relying on the loss value only. But we compare the visualized result using PCA dimension reduction method at this stage. Note that we do not choose T-SNE, because using linear dimension reduction method can intuitively better reflect how similar the generated data are. We observe that the Generator can keep generating diverse data points and start to provide better data points coverage against original load pattern data. This observation suggests that the Generator is slowly learning the load pattern and distributions from the training data. It can also be well reflected by the Fr\'echet Distance as the value keeps decreasing. After the training process reaches a certain point, the discriminator starts to dominate the training process where the generator loss starts to increase significantly. As shown in Figure \ref{model_selection}, if we use the Generator weight at this stage, regardless of the input $z$, similar load pattern data will be generated and the mode collapse problem occurred. The Fr\'echet Distance curve also shows an increase at this certain point, demonstrating its ability to reflect mode collapse.

\subsection{Benchmark Result Evaluation}
RLP-GAN can generate synthetic residential load patterns with well-retained distribution and temporal feature correlation against original data, while still being able to capture the diverse patterns from real-life load data. Demonstrated by both sequence-level and aggregated-level evaluations mentioned in Section \ref{subsubsec:Sequence_Level_Evaluation} and Section \ref{subsubsec:Aggregated_Evaluation} against the other four benchmark models, RLP-GAN has achieved a considerable amount of performance gain. Through further analysis of the evaluation results, we obtain the following four key findings.

\subsubsection{All five generative models (i.e. RLP-GAN, WGAN, ACGAN, C-RNN-GAN, DDPM) are able to capture certain electricity load patterns from original data with a well-retained correlation between different time steps}
\begin{figure*}
    \centering
    \begin{minipage}{0.245\textwidth}
        \centering
        \includegraphics[width=\linewidth]{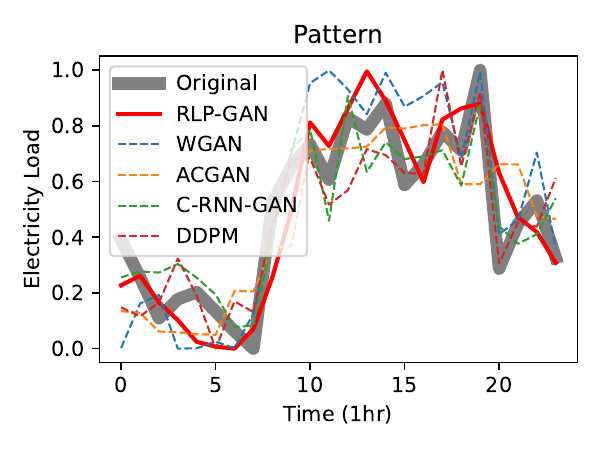}
    \end{minipage}
    \begin{minipage}{0.245\textwidth}
        \centering
        \includegraphics[width=\linewidth]{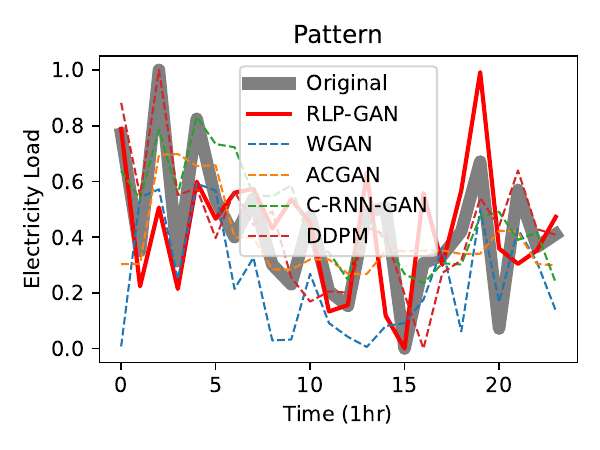}
    \end{minipage}
    \begin{minipage}{0.245\textwidth}
        \centering
        \includegraphics[width=\linewidth]{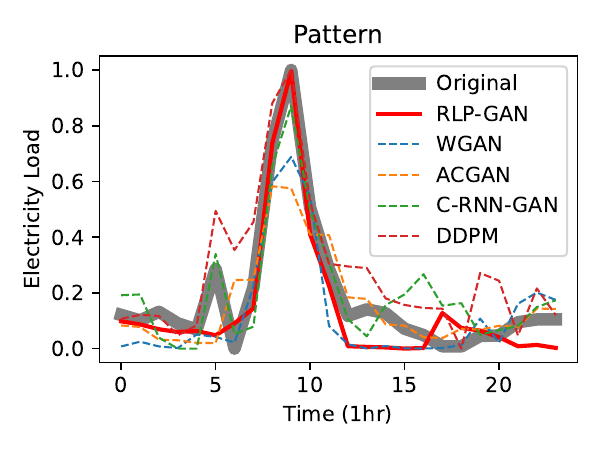}
    \end{minipage}
    \begin{minipage}{0.245\textwidth}
        \centering
        \includegraphics[width=\linewidth]{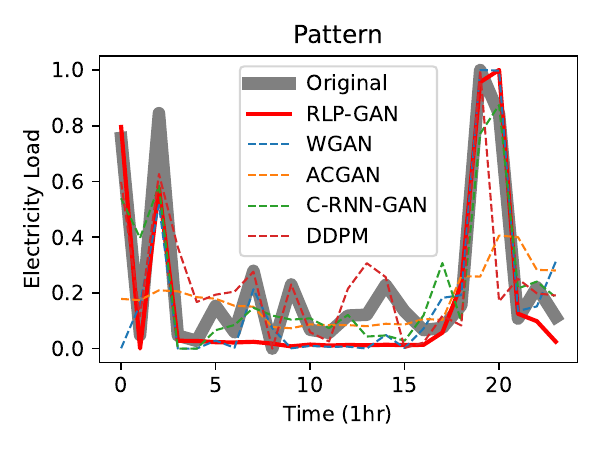}
    \end{minipage}

    \begin{minipage}{0.245\textwidth}
        \centering
        \includegraphics[width=\linewidth]{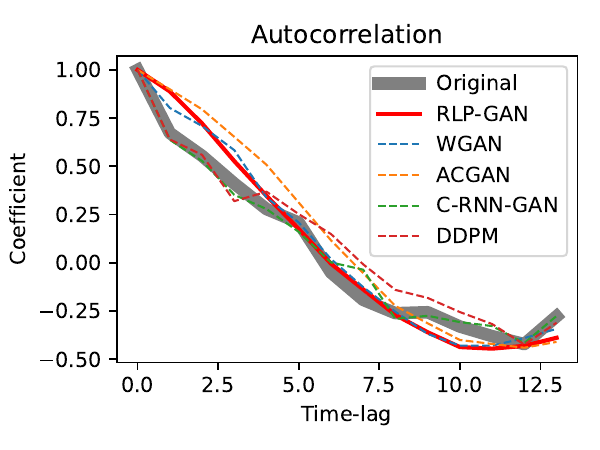}
    \end{minipage}
    \begin{minipage}{0.245\textwidth}
        \centering
        \includegraphics[width=\linewidth]{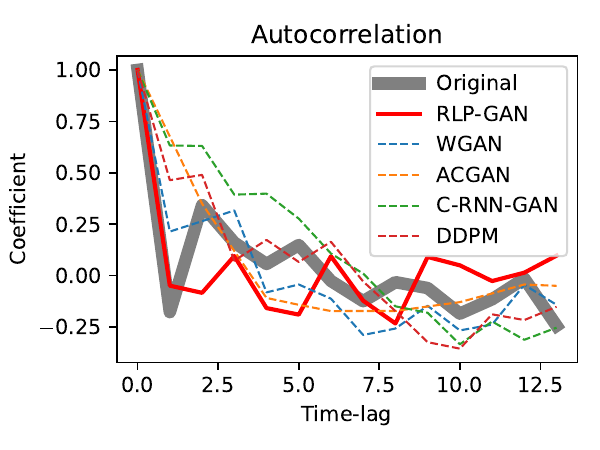}
    \end{minipage}
    \begin{minipage}{0.245\textwidth}
        \centering
        \includegraphics[width=\linewidth]{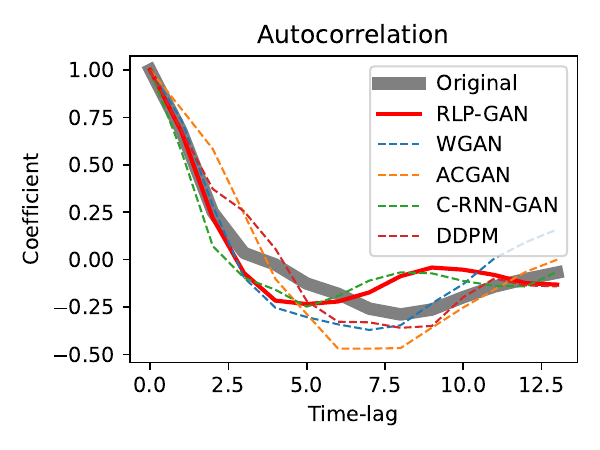}
    \end{minipage}
    \begin{minipage}{0.245\textwidth}
        \centering
        \includegraphics[width=\linewidth]{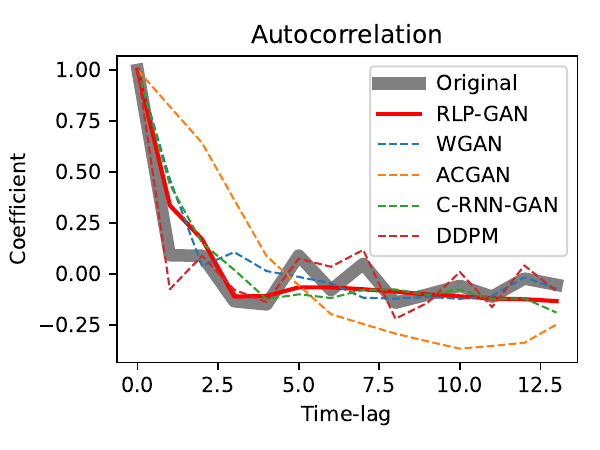}
    \end{minipage}
\caption{Similarity comparison of selected samples from our testing dataset against samples from generated data. The sample pairs are selected by calculating the Euclidean distance and select the minimum one. We also provide the auto-correlations plots of selected sample pairs to verify generated samples' ability to capture time-correlations.}
\label{gen_sequence_overview}
\end{figure*}

As shown in Figure \ref{gen_sequence_overview}, we randomly select four generated data samples from RLP-GAN and four benchmark models. Based on each selected synthetic data sample, we look for the original samples with minimum Euclidean distance from the testing dataset and visualize them along with their auto-correlation. Figure \ref{gen_sequence_overview} shows the visualization result of RLP-GAN.  We evaluate data generated by four benchmark models by following the same way. We observe that all models are able to capture some load patterns from the original dataset while retaining correlation between different time steps. \textit{However, based on the results from other evaluation methods, this does not suggest all models are able to generate datasets with similar distributions and diverse load patterns.}

\subsubsection{Pre-clustering methods can affect the overall diversity of generated electricity load patterns}
As shown in Figure \ref{dimension_reduction_visualization}, RLP-GAN demonstrates leading performance in terms of diversity and similarity of generated load pattern data. Although the pre-clustering method is proposed to further ensure the diversity of generated data, our PCA-based visualization plot shows that in terms of the whole dataset, the pre-clustering method can cause the generated data too close to the centers of different clusters. This is getting even more obvious when we visualize using a more powerful non-linear dimension reduction method (i.e. T-SNE) based on Stochastic Neighbor Embedding. Even though each clustering center is well separated, the generated data are too close to the clustering center while there is no coverage between two or more clustering centers. This can also be attributed to the fact that ACGAN model is a variant of conditional GAN model. Its discriminator is not only responsible for distinguishing real or fake data, but it is also used to classify intake data. The discriminator can guide the generator to generate data that are easier to classify, thus the generated data can be close to the clustering centers.

\subsubsection{Compared to CNN-based models, LSTM-based models can generate more diverse load pattern data by effectively capturing temporal dependencies and modeling more complex electricity load pattern data}
LSTM-based models are better suited for temporal or sequential data like load data due to their strong ability of modeling complex long-term dependencies. Our experiment results in the bottom of Figure \ref{dimension_reduction_visualization} with the T-SNE dimension reduction method also demonstrate that the data generated by two LSTM-based models (i.e. RLP-GAN and C-RNN-GAN) achieve significantly better coverage of original load patterns compared to the other three CNN-based models (i.e. WGAN, ACGAN and DDPM). 
From the aggregated evaluation, we observe that the CNN-based DDPM performs the worst in the CDF plot, as shown in Figure \ref{CDF}, highlighting the limitations of CNN-based approaches in this task. Among other models, the performance difference between the RNN-based RLP-GAN and C-RNN-GAN, as well as the CNN-based WGAN and ACGAN, is subtle in the CDF plot. However, the RNN-based models (RLP-GAN and C-RNN-GAN) demonstrate a clear advantage in both distance metrics, as shown in Table \ref{tab:Distance_Measurement}, and statistical measurements, as shown in Table \ref{tab:Statistical_Measurement}, suggesting that RNN architectures are better suited for capturing the temporal dependencies in this task compared to CNN-based architectures.

\subsubsection{RLP-GAN with LSTM-based over-complete autoencoder and weakly-supervised training process can further improve the model performance by accurately learning the sequence level temporal dependencies}
\begin{figure*}[!ht]
\centering
\includegraphics[width=\textwidth]{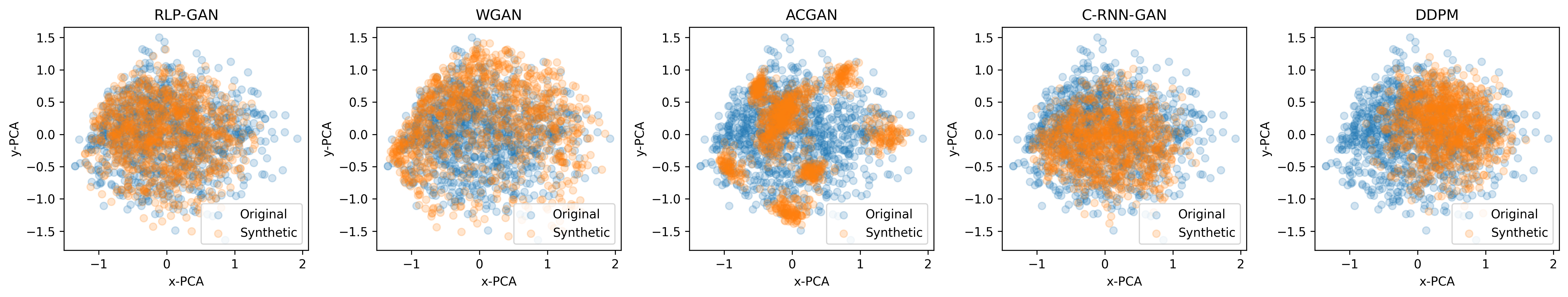} \includegraphics[width=\textwidth]{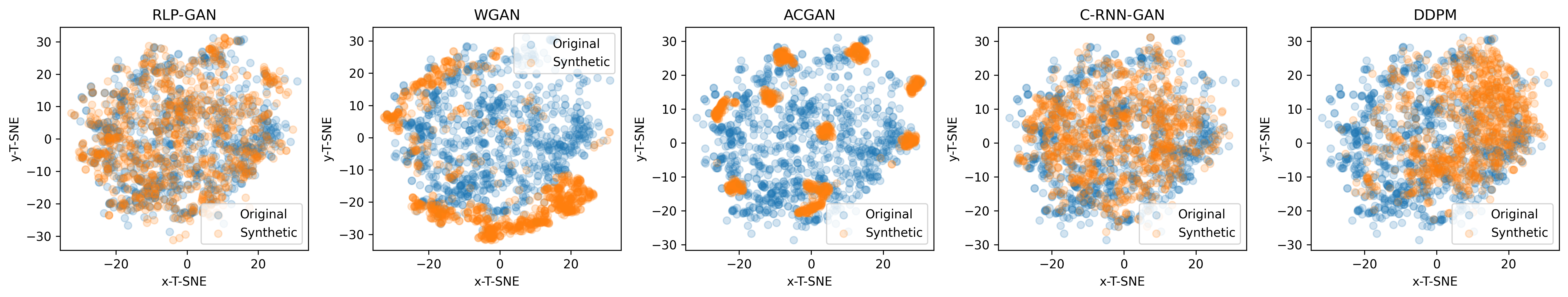}
\caption{Similarity and Diversity comparison using high dimensional load data visualization technique by applying dimension reduction methods to compare synthetic data against original data. Synthetic data are generated by RLP-GAN and four benchmark models displayed in four separate plots. The top five plots are using PCA to perform dimension reduction, and the bottom five plots are based on T-SNE to perform dimension reduction.}
\label{dimension_reduction_visualization}
\end{figure*}

\begin{figure}[h!]
\centering
\includegraphics[width=\columnwidth]{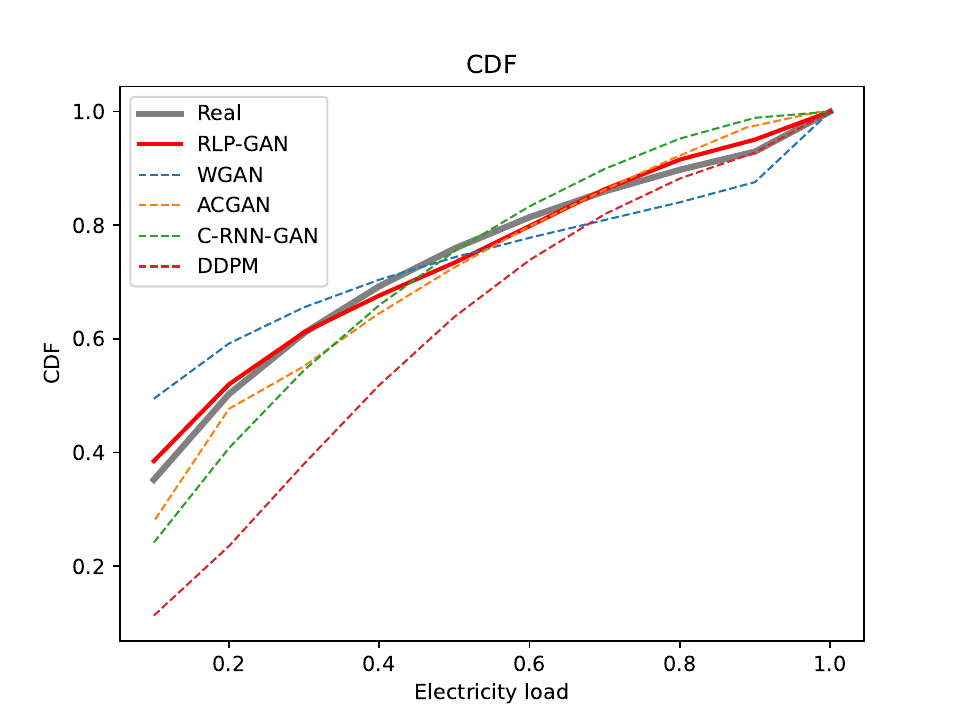}
\caption{Similarity comparison of Cumulative Distribution Function (CDF) of aggregate time-series sequences from original and synthetic datasets. Synthetic datasets are generated using RLP-GAN and four benchmark models aiming to capture similar CDF patterns as the original dataset.}
\label{CDF}
\end{figure}

The hidden space of LSTM-based over-complete autoencoders should provide a better representation of structures and patterns inherent in the input data, this can improve the capability of RLP-GAN to capture more complex load pattern data. On the other hand, the weakly-supervised training approach can further push the Generator of RLP-GAN to more accurately learn the temporal dependencies from the training data. Our evaluation results  demonstrate the effectiveness of these two methods mentioned. The RLP-GAN achieves the best overall load pattern coverage against four benchmark models, as shown in Figure \ref{dimension_reduction_visualization}. In Table \ref{tab:Distance_Measurement}, RLP-GAN can generate a dataset with the lowest distance against the test dataset measured by all the metrics, compared to the other four benchmark models. Based on statistical measurements shown in Table \ref{tab:Statistical_Measurement}, the synthetic dataset generated by RLP-GAN achieves the closet value against the test dataset in almost all measurements despite slightly being beaten by C-RNN-GAN in Qrt.3. In the aggregated dataset CDF visualization in Figure \ref{CDF}, synthetic dataset generated by RLP-GAN can also best approximate the CDF of testing dataset, compared to other four benchmark models.

\begin{table}[]
\centering
\caption{Result of Distance Measurements.}
\label{tab:Distance_Measurement}
\resizebox{\columnwidth}{!}{%
\begin{tabular}{|c|ccccc|}
\hline
{\begin{tabular}[c]{@{}c@{}}Distance\\ Measuremen\end{tabular}} &
  { \textbf{RLP-GAN}} &
  { WGAN} &
  { ACGAN} &
  { C-RNN-GAN} &
  { DDPM}\\ \hline
J-S Distance &
  \textbf{0.00770} &
  0.18363 &
  0.05576 &
  0.04277 &
  0.04451\\
RMSE &
  \textbf{0.03522} &
  0.56700 &
  0.26113 &
  0.19247 &
  0.59255\\ 
  Forward KL &
  \textbf{0.00026} &
  0.23141 &
  0.01286 &
  0.00731 &
  0.00780\\
  Reverse KL &
  \textbf{0.00026} &
  0.11331 &
  0.01214 &
  0.00738 &
  0.00809\\
  \hline
\end{tabular}%
}
\end{table}

\begin{table}[]
\centering
\caption{Result of Statistical Measurements.}
\label{tab:Statistical_Measurement}
\resizebox{\columnwidth}{!}{%
\begin{tabular}{|c|cccccc|}
\hline
{\color[HTML]{000000} \begin{tabular}[c]{@{}c@{}}Statistical\\ Measurement\end{tabular}} &
  {\color[HTML]{000000} \textbf{Original}} &
  {\color[HTML]{000000} \textbf{RLP-GAN}} &
  {\color[HTML]{000000} ACGAN} &
  {\color[HTML]{000000} WGAN} &
  C-RNN-GAN & DDPM\\ \hline
Mean  & \textbf{0.303} & \textbf{0.302} & 0.289 & 0.325 & 0.319 & 0.424\\ \hline
Std   & \textbf{0.300} & \textbf{0.296} & 0.350 & 0.271 & 0.249 & 0.273\\ \hline
Min   & \textbf{0.000} & \textbf{0.000} & 0.000 & 0.003 & 0.000 & 0.000\\ \hline
Max   & \textbf{1.000} & \textbf{1.000} & 1.000 & 0.983 & 1.000 & 1.000\\ \hline
Qrt.1 & \textbf{0.052} & \textbf{0.041} & 0.008 & 0.091 & 0.105 & 0.210\\ \hline
Qrt.2 & \textbf{0.198} & \textbf{0.190} & 0.104 & 0.228 & 0.266 & 0.386\\ \hline
Qrt.3 & \textbf{0.485} & \textbf{0.506} & 0.517 & 0.529 & 0.496 & 0.613\\ \hline
\end{tabular}%
}
\end{table}

\section{Discussions of Practical Implications and Future Directions}

\subsection{Practical Integration of RLP-GAN}

RLP-GAN's synthetic load pattern data has significant potential to enhance energy management systems (EMS). By augmenting historical load data with synthetic load patterns that capture a broader range of temporal dynamics, load forecasting models can learn richer representations of consumption behaviors and achieve improved predictive performance. In addition, incorporating synthetic data into EMS simulation tools enables a more comprehensive evaluation of demand response strategies, as the enriched dataset can better mimic real-world variations than historical data alone.

Such integration of RLP-GAN also requires practical considerations, such as alignment of data formats and the development of robust data pipelines to ensure seamless information flow into forecasting models and simulation platforms. Regulatory sides also need to be considered, including data integrity, cybersecurity compliance, and adherence to standardization requirements to safeguard the operational environment. Although synthetic data typically present lower privacy risks compared to raw consumption data, strict management protocols are necessary to prevent any potential re-identification or misuse of the generated information.

\subsection{Future Directions}
Despite the strong performance of RLP-GAN in generating realistic and diverse load patterns, several opportunities exist for further refinement to address other aspects of challenges. One promising direction is adapting the model to generate load patterns in different regions through fine-tuning or transfer learning strategies. Such adaptations would enable the model to efficiently capture region-specific consumption behaviors with minimal computational cost, thereby enhancing the practical deployment of generation models across diverse geographical and cultural contexts.

Additionally, while the current framework implicitly captures atypical or rare load events through its focus on distributional fidelity and diversity, there remains a need for explicit mechanisms that can reliably detect and replicate these anomalies. Enhancing robustness of generation models to noisy data and extreme events would not only improve the fidelity of generated patterns but also extend its utility for practical energy resilience planning. These research directions will be critical in bridging the gap between high-fidelity synthetic data generation and its seamless integration into real-world systems, enhancing energy management, demand response, and grid operations.

\section{Conclusion}
Our work developed a comprehensive framework for generating synthetic residential load patterns, referred to as the RLP-GAN model. This GAN-based model leverages an over-complete autoencoder and weakly-supervised training to effectively capture diverse temporal dependencies during the training process. To address the inherent instability and the risk of mode collapse in the GAN training, we employ a Fr\'echet distance measurement to select effective model weights. To evaluate the accuracy and diversity of the generated data, we conducted both quantitative and qualitative analyses at both the sequence and aggregate levels. Our results demonstrated that the RLP-GAN model outperforms four benchmark models, including ACGAN, WGAN, C-RNN-GAN and DDPM, in terms of the similarity and diversity of the generated data. In addition, we also discussed practical strategies for integrating the generated data into energy management systems and outlined future research directions aimed at further enhancing the model's adaptability and robustness.

\bibliographystyle{IEEEtran}
\bibliography{references}

\vfill

\begin{IEEEbiography}
[{\includegraphics[width=1in,height=1.25in,clip,keepaspectratio]{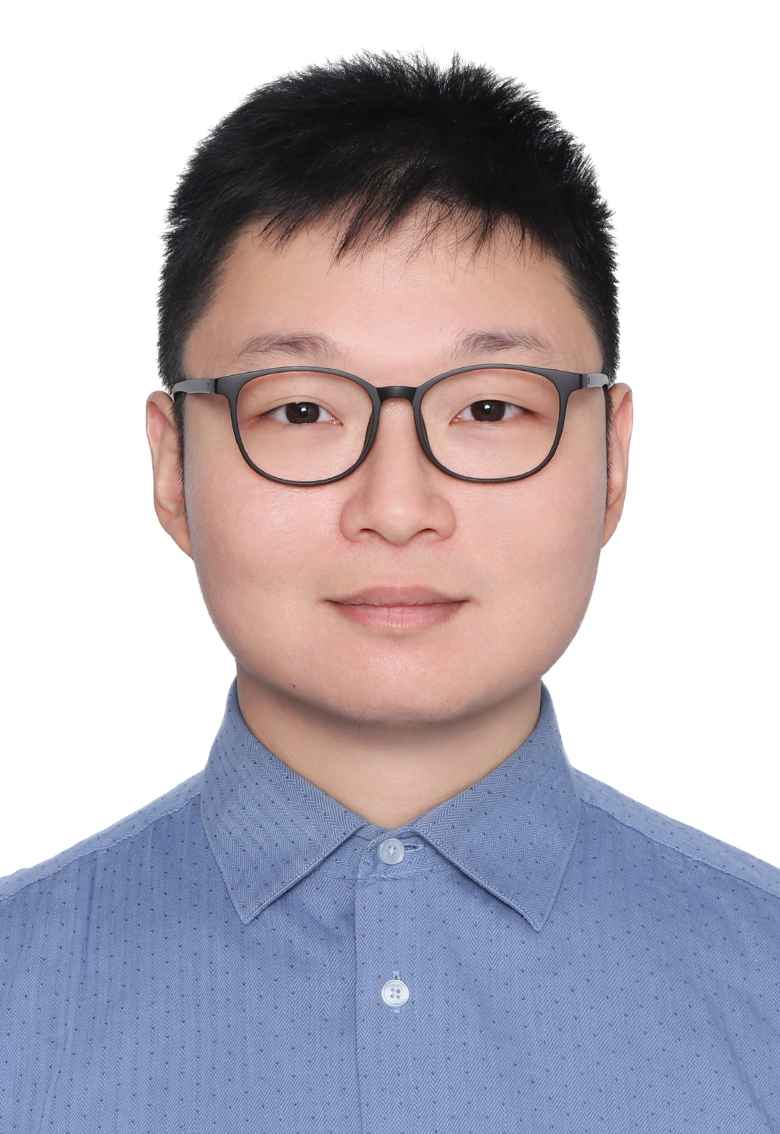}}]{Xinyu Liang} received his Bachelor of Computer Science and Master of Artificial Intelligence degrees from Monash University, Melbourne, VIC, Australia, in 2018 and 2021, respectively. He is currently pursuing his Ph.D. degree in Information Technology in the Department of Data Science and AI, Faculty of Information Technology, Monash University, Melbourne, VIC, Australia. His research interests include energy informatics and human-in-the-loop AI for smart energy systems.
\end{IEEEbiography}

\begin{IEEEbiography}
[{\includegraphics[width=1in,height=1.25in,clip,keepaspectratio]{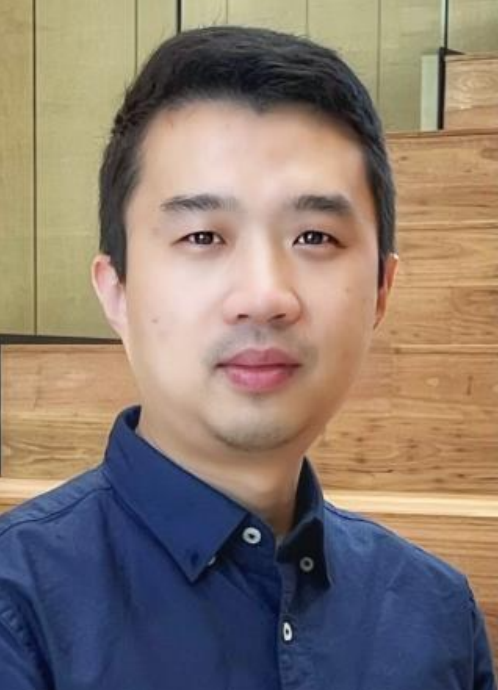}}]{Hao Wang} (M'16) received his Ph.D. in Information Engineering from The Chinese University of Hong Kong, Hong Kong, in 2016. He was a Postdoctoral Research Fellow at Stanford University, Stanford, CA, USA, and a Washington Research Foundation Innovation Fellow at the University of Washington, Seattle, WA, USA. He is currently a Senior Lecturer and ARC DECRA Fellow in the Department of Data Science and AI, Faculty of IT, Monash University, Melbourne, VIC, Australia. His research interests include optimization, machine learning, and data analytics for power and energy systems.
\end{IEEEbiography}

\end{document}